 \documentclass[journal]{IEEEtran}
\usepackage[hidelinks]{hyperref}
\usepackage{caption}
\usepackage{algorithm}
\usepackage{algorithmic}
\usepackage{graphicx}
\usepackage{ulem}
\usepackage{enumerate}
\usepackage{amsthm,amsmath}
\usepackage{epstopdf}
\usepackage{amsmath,amssymb,epsfig,slidesec,colortbl}
\usepackage[absolute,overlay]{textpos}
\usepackage{multimedia,algorithm,algorithmic,multirow,url}
\usepackage{cite}
\floatname{algorithm}{Algorithm 1}
\usepackage{bbm}
\usepackage{subfig}
\usepackage{color}
\usepackage{url}
\usepackage{enumitem}
\usepackage{balance}
\usepackage{tcolorbox}

\usepackage{tikz}
\usetikzlibrary{positioning,shapes.geometric,arrows.meta,calc,decorations.pathreplacing,fit}

\usepackage[font={small}]{caption}

\newtheorem{Obs}{Observation}
\newtheorem{Question}{Question}

\usepackage{geometry}
\IEEEoverridecommandlockouts
\begin{document}

\title{Decoupled Split Learning via Auxiliary Loss}
\author{Anower Zihad$^*$, Felix Owino$^*$, Ming Tang, and Chao Huang
\thanks{$*$ Equal contribution.}
\thanks{Anower Zihad, Felix Owino, and Chao Huang are with the School of Computing, Montclair State University, New Jersey, USA; Emails: \{zihada1,owinof1,huangch\}@montclair.edu. Ming Tang is with the Department of Computer Science and Engineering, Southern University of Science and Technology, Shenzhen, China; Email: tangm3@sustech.edu.cn.}
}

%

\vspace{-10mm}
\maketitle
\begin{abstract}
Split learning is a distributed training paradigm where a neural network is partitioned between clients and a server, which allows data to remain at the client while only intermediate activations are shared. Traditional split learning relies on end-to-end backpropagation across the client–server split point. This incurs a large communication overhead (i.e., forward activations and backward gradients need to be exchanged every iteration) and significant memory use (for storing activations and gradients). In this paper, we develop a beyond-backpropagation training method for split learning. In this approach, the client and server train their model partitions semi-independently, using local loss signals instead of propagated gradients. In particular, the client’s network is augmented with a small auxiliary classifier at the split point to provide a local error signal, while the server trains on the client’s transmitted activations using the true loss function. This decoupling removes the need to send backward gradients, which cuts communication costs roughly in half and also reduces memory overhead (as each side only stores local activations for its own backward pass). We evaluate our approach on CIFAR-10 and CIFAR-100. Our experiments show two key results. First, the proposed approach achieves performance on par with standard split learning that uses backpropagation. Second, it  significantly reduces communication (of transmitting activations/gradient) by $50\%$ and peak memory usage by up to $58\%$. 
\end{abstract}

\section{Introduction}
In distributed machine learning, multiple parties collaboratively train a model across different devices or institutions without centralizing all the training data. One example is \textbf{split learning (SL)}, where a neural network is partitioned across clients and a server so that raw data never leaves the local devices and only intermediate activations and gradients are communicated during training \cite{gupta2018distributed}. SL is beneficial in resource-constrained environments as each client only trains a portion of the neural network locally, which reduces their computational burden by handling the initial layers of the model while the rest is processed on the server.

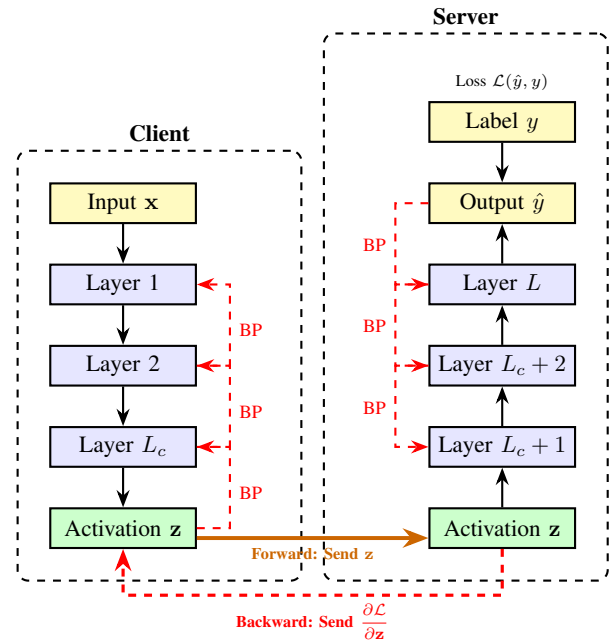
\begin{figure}[t]
\centering
\begin{tikzpicture}[
    node distance=0.6cm and 1.5cm,
    scale=0.88,
    every node/.style={transform shape},
    box/.style={rectangle, draw, thick, minimum width=2.2cm, minimum height=0.6cm, align=center, fill=blue!10, font=\normalfont},
    databox/.style={rectangle, draw, thick, minimum width=2.2cm, minimum height=0.6cm, align=center, fill=yellow!30, font=\normalfont},
    activation/.style={rectangle, draw, thick, minimum width=2.2cm, minimum height=0.6cm, align=center, fill=green!20, font=\normalfont},
    arrow/.style={-Stealth, thick},
    redarrow/.style={-Stealth, thick, red, dashed},
    orangearrow/.style={-Stealth, ultra thick, orange!80!black},
    section/.style={draw, thick, dashed, rounded corners, inner sep=12pt}
]

\node[databox] (input) {Input $\mathbf{x}$};
\node[box, below=of input] (layer1) {Layer 1};
\node[box, below=of layer1] (layer2) {Layer 2};
\node[box, below=of layer2] (layer3) {Layer $L_c$};
\node[activation, below=of layer3] (cutlayer) {Activation $\mathbf{z}$};

\node[activation, right=3.5cm of cutlayer] (server_input) {Activation $\mathbf{z}$};
\node[box, above=of server_input] (layer4) {Layer $L_c+1$};
\node[box, above=of layer4] (layer5) {Layer $L_c+2$};
\node[box, above=of layer5] (layer6) {Layer $L$};
\node[databox, above=of layer6] (output) {Output $\hat{y}$};

\node[databox, above=0.6cm of output] (label) {Label $y$};

\node[above=0.05cm of label, font=\scriptsize\normalfont] (loss_label) {Loss $\mathcal{L}(\hat{y}, y)$};

\draw[arrow] (input) -- (layer1);
\draw[arrow] (layer1) -- (layer2);
\draw[arrow] (layer2) -- (layer3);
\draw[arrow] (layer3) -- (cutlayer);

\draw[redarrow] (cutlayer.east) -- ++(0.5,0) coordinate (bp1start) |- (layer3.east);
\draw[redarrow] (layer3.east) -- ++(0.5,0) coordinate (bp2start) |- (layer2.east);
\draw[redarrow] (layer2.east) -- ++(0.5,0) coordinate (bp3start) |- (layer1.east);

\node[anchor=south west, font=\footnotesize\normalfont, red, xshift=0.02cm, yshift=0.32cm] at (bp1start) {BP};
\node[anchor=south west, font=\footnotesize\normalfont, red, xshift=0.02cm, yshift=0.32cm] at (bp2start) {BP};
\node[anchor=south west, font=\footnotesize\normalfont, red, xshift=0.02cm, yshift=0.32cm] at (bp3start) {BP};

\coordinate (client_right) at ($(bp3start) + (0.6,0)$);
\node[section, fit=(input) (layer1) (layer2) (layer3) (cutlayer) (client_right), label={[font=\normalfont\bfseries]above:Client}] (client_box) {};

\draw[arrow] (server_input) -- (layer4);
\draw[arrow] (layer4) -- (layer5);
\draw[arrow] (layer5) -- (layer6);
\draw[arrow] (layer6) -- (output);
\draw[arrow] (label) -- (output);

\draw[redarrow] (output.west) -- ++(-0.5,0) coordinate (bp6start) |- (layer6.west);
\draw[redarrow] (layer6.west) -- ++(-0.5,0) coordinate (bp5start) |- (layer5.west);
\draw[redarrow] (layer5.west) -- ++(-0.5,0) coordinate (bp4start) |- (layer4.west);

\node[anchor=south east, font=\footnotesize\normalfont, red, xshift=-0.02cm, yshift=-0.85cm] at (bp6start) {BP};
\node[anchor=south east, font=\footnotesize\normalfont, red, xshift=-0.02cm, yshift=-0.85cm] at (bp5start) {BP};
\node[anchor=south east, font=\footnotesize\normalfont, red, xshift=-0.02cm, yshift=-0.85cm] at (bp4start) {BP};

\coordinate (server_left) at ($(bp6start) + (-0.6,0)$);
\node[section, fit=(server_input) (layer4) (layer5) (layer6) (output) (label) (loss_label) (server_left), label={[font=\normalfont\bfseries]above:Server}] (server_box) {};

\coordinate (forward_start) at ($(cutlayer.east) + (0,-0.15)$);
\coordinate (forward_end) at ($(server_input.west) + (0,-0.15)$);
\draw[orangearrow, line width=1.5pt] 
    (forward_start) -- (forward_end)
    node[midway, below, font=\scriptsize\normalfont\bfseries, orange!80!black] {Forward: Send $\mathbf{z}$};

\draw[red, dashed, line width=1.2pt, -Stealth] 
    (server_input.south) -- ++(0,-0.7) -| (cutlayer.south);

\node[font=\scriptsize\normalfont\bfseries, red] at ($(cutlayer.south)!0.5!(server_input.south) + (0,-1.1)$) 
    {Backward: Send $\displaystyle\frac{\partial \mathcal{L}}{\partial \mathbf{z}}$};



\end{tikzpicture}
\caption{A typical SL architecture showing forward propagation of activations (orange arrow) and back propagation (BP) of gradients (red dashed arrow) between a client and the server. Bi-directional communications are required per training iteration.}
\label{fig:traditional_split}
 \vspace{-2mm}
\end{figure}

More specifically, a canonical SL setup involves splitting a neural network (in total $L$ layers) into two parts: a client-side model (from layer $1$ to $L_c$) on the client side and a server-side model (from layer $L_c+1$ to $L$) on the server side. Layer $L_c$ in which the network is split is called the cut layer (see also Fig.~\ref{fig:traditional_split}). During a forward pass, the client computes intermediate activations and sends them (and the labels) to the server.\footnote{Sharing labels but not raw features is widely considered in SL. One can adopt a U-shaped SL structure that does not share labels either \cite{gupta2018distributed}.} The server continues the forward propagation to compute predictions. In the backward pass, the server evaluates the loss and sends the gradient of the intermediate activation back to the client, which then updates the clients-side model’s parameters using the chain rule. This approach keeps data local and only exchanges intermediate features and gradients, which helps preserve privacy. It also enables less heavy client-side computation (compared to other distributed learning approaches such as federated learning), since the clients only need to train part of the model.

However, conventional SL still relies on backpropagation (\textbf{BP}) across the network split.
which means each training step requires two rounds of communication (i.e., activations forward, gradients backward) and that the client must wait for the server’s gradient to update its parameters. 
For large models or high-dimensional layer cuts, these gradients can incur a significant \underline{communication cost}. In addition, both sides need to store activations for use in backward propagation, which could contribute to a large \underline{memory usage}. It is crucial to reduce these overheads for scalable and resource-efficient distributed learning, especially when clients are edge devices with limited bandwidth and memory \cite{ilhan2023scalefl}.

In this paper, we develop a beyond-BP approach that addresses these challenges by decoupling the training process at the split point. Instead of using BP signals between the server and clients, our method allows each side of the split network to be trained with its own local learning objective. The major inspiration comes from \cite{belilovsky2020decoupled}, which demonstrated that deep networks can be trained in a layer-wise greedy fashion using local losses at intermediate layers. However, it focuses a centralized learning setting, and hence did not consider how model split in SL affects model performance and overheads in a distributed setting. 

To fill this gap, we design a beyond-BP approach tailored for SL, called decoupled SL, which attaches a lightweight auxiliary classifier at the cut layer on the client’s side to generate a local error signal for the client-side model. The server, on the other hand, continues to train its server-side model using the end task loss (e.g., classification loss) computed with the  labels. Importantly, the client’s update no longer depends on receiving the server’s gradient. Instead, it can compute its own gradient immediately after the forward pass using the auxiliary loss. The server’s update is performed in parallel using the activations received from the client. 
As a result, the training process requires only a single round of communication per iteration (sending intermediate activations) and no backward transmission, which substantially improves communication efficiency. Memory usage is also reduced because each side only needs to store intermediate results for its own backward pass and can discard them once local updates are done, without the need to hold data for a backpropagation step.

We aim to answer two research questions in this work:
\begin{Question}
Can our decoupled split learning approach maintain comparable performance to BP-based split learning?
\end{Question}
\begin{Question}
How much is the overhead saving in terms of communication and memory usage?
\end{Question}

To answer \textbf{Q1}, we measure the convergence behavior of the proposed method versus standard SL on two public datasets. To answer \textbf{Q2}, we quantify the communication volume and memory footprint for each approach. Our contributions are summarized as follows.

\begin{itemize}
\item We develop a decoupled SL approach that removes cross-party gradient dependency by training the client-side and server-side models with separate local objectives at the split point. In particular, we design a lightweight auxiliary classifier at the cut layer that enables clients to update their local models immediately after the forward pass.
\item Experiments on CIFAR-10 and CIFAR-100 show that our approach significantly reduces communication and memory costs while maintaining model performance comparable to conventional BP-based SL.
\end{itemize}

The remainder of this paper is organized as follows. Section \ref{sec: related}
 reviews related work. Section \ref{sec: algorithm} details our decoupled SL approach. Section \ref{sec: experiments} presents experimental design and results. Section \ref{sec: conclusion} concludes the paper.

\section{Related Work}\label{sec: related}
\subsection{Split Learning}
SL was introduced as a privacy-preserving distributed training scheme that allows clients to jointly train a model without sharing raw data. In the seminal work \cite{gupta2018distributed}, the model is split between a client and server, and only the intermediate activations and gradients are exchanged. Subsequent research has applied SL to various domains, including healthcare, Internet of things, and graph data. A recent study \cite{hu2025review} provides a comprehensive review of SL variations and their performance trade-offs. One key observation is that, while SL can reduce data exposure, it still incurs network communication each training step and depends on synchronous end-to-end gradient exchanges. Hybrid schemes such as split federated learning \cite{thapa2022splitfed,han2024convergence} attempt to improve scalability, but none remove the fundamental need for backpropagating gradients between the clients and the sever. \textit{Our work differs in that we remove BP across the cut layer using decoupled training, which will be shown to substantially reduce communication and memory usage.} 

\subsection{Decoupled Training of Neural Networks}
The idea of decoupling the training neural networks has appeared in various forms. In large-scale distributed training of deep networks, pipeline parallelism \cite{narayanan2019pipedream} overlaps the computation of different micro-batches across layers to mitigate communication, but it still uses end-to-end BP. Feedback alignment \cite{nokland2016direct} alters the BP process by using random or direct error signals. While feedback alignment and it variants allow simultaneous layer updates, they only fail to match the performance of canonical BP on large tasks. 

There has been a resurgence of interest in greedy layer-wise training of neural networks, e.g., \cite{ye2025learning, belilovsky2019greedy, belilovsky2019greedy}. \cite{jaderberg2017decoupled} proposed using synthetic gradients, where an auxiliary network predicts layer-wise gradients from activations to remove backward dependency on BP. However, it does not scale well to complex architectures as predicting high-dimensional gradients leads to error accumulation and degraded accuracy.
\cite{belilovsky2019greedy} first showed that one can train each layer of a CNN sequentially with a local classifier and  achieve competitive accuracy on ImageNet. Building on this, DGL in \cite{belilovsky2020decoupled} enables training multiple network modules in parallel using local losses. 
However, these methods were mainly developed for centralized learning, which cannot be directly applied to distributed learning that requires careful workload balance between clients and the server.

Among the most relevant studies in distributed settings pertain to \cite{han2021accelerating, nair2025fsl}. In particular, \cite{han2021accelerating} used locally generated losses to accelerate federated learning using model split, but their method does not eliminate cross-split gradient dependency. \cite{nair2025fsl}  reduces latency by estimating smashed activation, yet it relies on gradient approximation and backward dependency across the split. \textit{In contrast, our approach removes BP across the cut layer by introducing an auxiliary loss, which enables decoupled updates with one-way activation communication and reduced communication and memory overhead.}

\section{Decoupled Split Learning Approach}\label{sec: algorithm}
\subsection{Overview}
In this section, we formalize the decoupled split learning method. For ease of presentation, we consider a two-party setup with a client and a server training a neural network collaboratively.\footnote{Our experiments will study both the one-client-one-server and multiple-client-one-server settings.} The network is split at the cut layer $L_c$.  Layers $[1, \dots, L_c]$ (the bottom client-side model $M_b$) reside on the client, and layers $[L_c+1, \dots, L]$ (the top server-side model $M_t$) reside on the server. For a given input $\mathbf{x}$ and label $y$, let $\mathbf{z} = M_b(\mathbf{x})$ be the intermediate activation at the cut layer. The server produces prediction $\hat{y} = M_t(\mathbf{z})$. In conventional SL, the training objective is the global loss $\mathcal{L}(\hat{y}, y)$ (e.g. cross-entropy for classification). Training updates for $M_b$ and $M_t$ are derived from the gradient of $\mathcal{L}$: $\frac{\partial \mathcal{L}}{\partial \theta_b}$ and $\frac{\partial \mathcal{L}}{\partial \theta_t}$ obtained via BP through the entire network, where $\theta_b$ and $\theta_t$ correspond to the model parameters at the client side and server side, respectively. Different from this, our approach modifies the training objective to enable decoupled optimization of $M_b$ and $M_t$. 

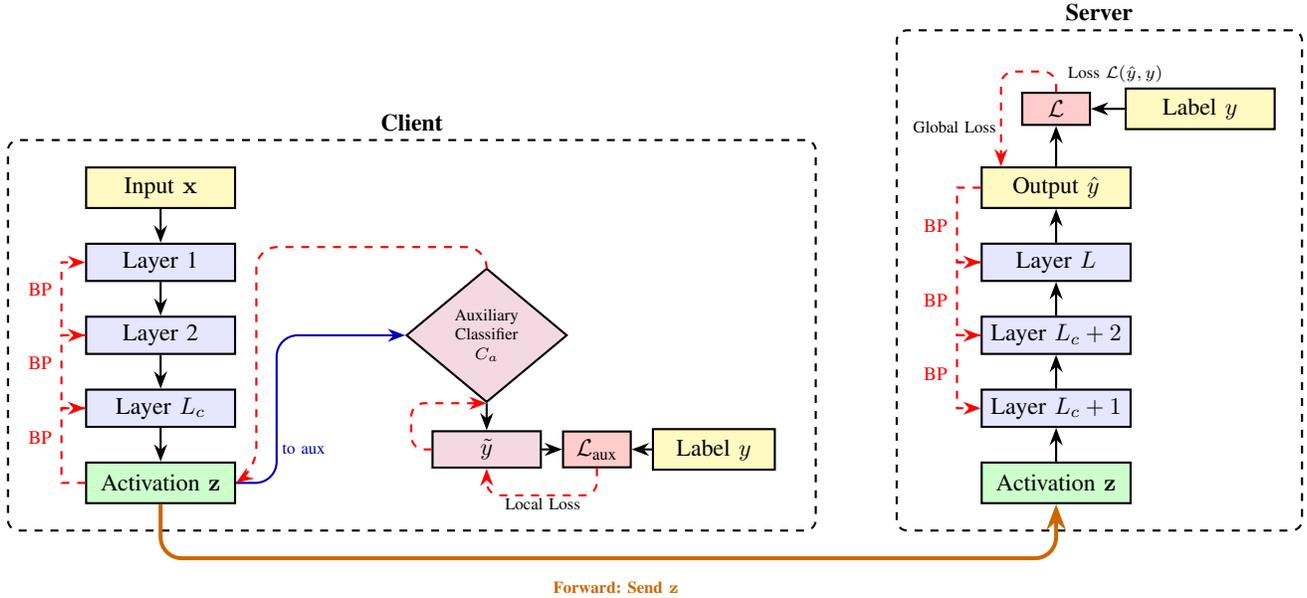
\begin{figure*}[t]
\centering
\begin{tikzpicture}[
    node distance=0.5cm and 1.5cm,
    scale=0.9,
    every node/.style={transform shape},
    box/.style={rectangle, draw, thick, minimum width=2.2cm, minimum height=0.5cm, align=center, fill=blue!10, font=\normalfont},
    databox/.style={rectangle, draw, thick, minimum width=2.2cm, minimum height=0.6cm, align=center, fill=yellow!30, font=\normalfont},
    activation/.style={rectangle, draw, thick, minimum width=2.2cm, minimum height=0.6cm, align=center, fill=green!20, font=\normalfont},
    auxbox/.style={rectangle, draw, thick, minimum width=1.6cm, minimum height=0.5cm, align=center, fill=purple!15, font=\normalfont},
    arrow/.style={-Stealth, thick},
    redarrow/.style={-Stealth, thick, red, dashed},
    bluearrow/.style={-Stealth, thick, blue!70!black},
    orangearrow/.style={-Stealth, ultra thick, orange!80!black},
    section/.style={draw, thick, dashed, rounded corners, inner sep=10pt},
    loss/.style={rectangle, draw, thick, minimum width=1.0cm, minimum height=0.45cm, align=center, fill=red!20, font=\normalfont}
]

\node[databox] (input) {Input $\mathbf{x}$};
\node[box, below=of input] (layer1) {Layer 1};
\node[box, below=of layer1] (layer2) {Layer 2};
\node[box, below=of layer2] (layer3) {Layer $L_c$};
\node[activation, below=of layer3] (cutlayer) {Activation $\mathbf{z}$};

\draw[arrow] (input) -- (layer1);
\draw[arrow] (layer1) -- (layer2);
\draw[arrow] (layer2) -- (layer3);
\draw[arrow] (layer3) -- (cutlayer);

\draw[redarrow] (cutlayer.west) -- ++(-0.35,0) coordinate (bp1start) |- (layer3.west);
\draw[redarrow] (layer3.west) -- ++(-0.35,0) coordinate (bp2start) |- (layer2.west);
\draw[redarrow] (layer2.west) -- ++(-0.35,0) coordinate (bp3start) |- (layer1.west);

\node[anchor=south east, font=\footnotesize\normalfont, red, xshift=-0.02cm, yshift=0.46cm] at (bp1start) {BP};
\node[anchor=south east, font=\footnotesize\normalfont, red, xshift=-0.02cm, yshift=0.46cm] at (bp2start) {BP};
\node[anchor=south east, font=\footnotesize\normalfont, red, xshift=-0.02cm, yshift=0.46cm] at (bp3start) {BP};

\node[diamond, draw, thick, minimum width=1.8cm, minimum height=1.8cm, align=center, fill=purple!15, font=\scriptsize\normalfont, aspect=1.2, right=2.5cm of layer2] (auxclass) {Auxiliary\\Classifier\\$C_a$};

\node[auxbox, below=0.4cm of auxclass] (auxpred) {$\tilde{y}$};
\node[loss, right=0.3cm of auxpred] (clientloss) {$\mathcal{L}_{\text{aux}}$};
\node[databox, right=0.3cm of clientloss, minimum width=1.8cm] (clientlabel) {Label $y$};

\draw[bluearrow, rounded corners=8pt] (cutlayer.east) -- ++(0.6,0) |- (auxclass.west);
\node[blue!70!black, font=\scriptsize\normalfont] at ($(cutlayer.east)+(1.0,0.5)$) {to aux};

\draw[arrow] (auxclass.south) -- (auxpred);
\draw[arrow] (auxpred) -- (clientloss);
\draw[arrow] (clientlabel) -- (clientloss);

\draw[redarrow, rounded corners=5pt] (clientloss.south) -- ++(0,-0.40) -| (auxpred.south);
\draw[redarrow, rounded corners=5pt] (auxpred.west) -- ++(-0.3,0) |- (auxclass.south);
\draw[redarrow, rounded corners=8pt] (auxclass.north) -- ++(0,0.3) -| ($(cutlayer.east)+(0.3,0.2)$) -- (cutlayer.east);

\node[below=0.32cm of clientloss, xshift=-0.8cm,font=\scriptsize\normalfont] {Local Loss};

\coordinate (client_left) at ($(bp3start) + (-0.4,0)$);
\coordinate (client_right) at ($(clientlabel.east) + (0.2,0)$);
\coordinate (client_top) at ($(clientlabel.north) + (0,0.2)$);
\node[section, fit=(input) (layer1) (layer2) (layer3) (cutlayer) (auxclass) (auxpred) (clientloss) (clientlabel) (client_left) (client_right) (client_top), label={[font=\normalfont\bfseries]above:Client}] (client_box) {};

\node[activation, right=11.0cm of cutlayer] (server_input) {Activation $\mathbf{z}$};
\node[box, above=of server_input] (layer4) {Layer $L_c+1$};
\node[box, above=of layer4] (layer5) {Layer $L_c+2$};
\node[box, above=of layer5] (layer6) {Layer $L$};
\node[databox, above=of layer6] (output) {Output $\hat{y}$};

\node[loss, above=0.6cm of output] (serverloss) {$\mathcal{L}$};
\node[databox, right=0.5cm of serverloss] (serverlabel) {Label $y$};

\node[above=0.2cm of serverloss, xshift=25,yshift=-5,font=\scriptsize\normalfont] (loss_label) {Loss $\mathcal{L}(\hat{y}, y)$};

\draw[arrow] (server_input) -- (layer4);
\draw[arrow] (layer4) -- (layer5);
\draw[arrow] (layer5) -- (layer6);
\draw[arrow] (layer6) -- (output);
\draw[arrow] (output) -- (serverloss);
\draw[arrow] (serverlabel) -- (serverloss);

\draw[redarrow, rounded corners=5pt] (serverloss.north) -- ++(0,0.3) -| ($(output.north west)+(0.3,0)$);
\draw[redarrow] (output.west) -- ++(-0.35,0) coordinate (bp6start) |- (layer6.west);
\draw[redarrow] (layer6.west) -- ++(-0.35,0) coordinate (bp5start) |- (layer5.west);
\draw[redarrow] (layer5.west) -- ++(-0.35,0) coordinate (bp4start) |- (layer4.west);

\node[anchor=south east, font=\footnotesize\normalfont, red, xshift=-0.02cm, yshift=-0.78cm] at (bp6start) {BP};
\node[anchor=south east, font=\footnotesize\normalfont, red, xshift=-0.02cm, yshift=-0.78cm] at (bp5start) {BP};
\node[anchor=south east, font=\footnotesize\normalfont, red, xshift=-0.02cm, yshift=-0.78cm] at (bp4start) {BP};

\coordinate (server_left) at ($(bp6start) + (-0.5,0)$);
\node[section, fit=(server_input) (layer4) (layer5) (layer6) (output) (serverloss) (serverlabel) (loss_label) (server_left), label={[font=\normalfont\bfseries]above:Server}] (server_box) {};

\node[left=-0.35cm of output, yshift=0.9cm, font=\scriptsize\normalfont] {Global Loss};

\draw[orangearrow, line width=1.5pt, rounded corners=8pt] 
    (cutlayer.south) -- ++(0,-0.8) -| (server_input.south)
    node[midway, below, font=\scriptsize\normalfont\bfseries, orange!80!black, yshift=-0.2cm,xshift=-6.5cm] {Forward: Send $\mathbf{z}$};



\end{tikzpicture}
\caption{An illustration of our decoupled SL approach. The client trains using local loss $\mathcal{L}_{\text{aux}}$ computed via auxiliary classifier $C_a$, while the server trains using global loss $\mathcal{L}$. Only forward activations are communicated (orange arrow) and no backward gradients are sent.}
\label{fig:decoupled_split}
 \vspace{-2mm}
\end{figure*}

\textbf{Auxiliary Loss for the Client}. We introduce an auxiliary classifier $C_a$ with parameters $\theta_a$ at the cut layer output (see also Fig.~\ref{fig:decoupled_split}). $C_a$ is a shallow neural network (for example, a small multi-layer perceptron) that takes $\mathbf{z}$ as input and produces an auxiliary prediction $\tilde{y} = C_a(\mathbf{z})$. We define a local loss on the client side, $\mathcal{L}_{\text{aux}}(\tilde{y}, y)$, which serves as the training objective for the client’s model $M_b$ (and the auxiliary head $C_a$) in lieu of the true global loss. In other words, the client seeks to minimize $\mathcal{L}_{\text{aux}}(C_a(M_b(\mathbf{x})), y)$, making $M_b(\mathbf{x})$ a good intermediate representation for predicting $y$ on its own. 
Note that $C_a$ is typically lightweight, and hence optimizing $\mathcal{L}_{\text{aux}}$ does not overly burden the client, and it provides a meaningful direction to update $M_b$’s weights without full gradient information from $M_t$.

\begin{algorithm}[t]
\caption{Decoupled Split Learning with Auxiliary Loss}
\begin{algorithmic}[1]\label{Decoupled_Split_Learning_Algorithm}
\REQUIRE Training dataset $\mathcal{D} = \{(\mathbf{x}_i, y_i)\}_{i=1}^N$, client model $M_b$, server model $M_t$, auxiliary classifier $C_a$, learning rates $\eta_b$, $\eta_t$, batch size $B$, number of epochs $E$
\ENSURE Trained models $M_b$, $M_t$, $C_a$

\FOR{epoch $e = 1$ to $E$}
    \FOR{each mini-batch $\mathcal{B} = \{(\mathbf{x}_i, y_i)\}_{i=1}^B \subset \mathcal{D}$}
        
        \STATE \textbf{// Client Side: Forward and Local Backward}
        \FOR{each sample $(\mathbf{x}_i, y_i) \in \mathcal{B}$}
            \STATE Compute activation: $\mathbf{z}_i = M_b(\mathbf{x}_i)$
            \STATE Compute auxiliary prediction: $\tilde{y}_i = C_a(\mathbf{z}_i)$
        \ENDFOR

        \STATE \textbf{Send} $\{\mathbf{z}_i\}_{i=1}^B$ to server
        \STATE Compute auxiliary loss: $L_{\text{aux}} = \frac{1}{B}\sum_{i=1}^B \mathcal{L}_{\text{aux}}(\tilde{y}_i, y_i)$
        \STATE Compute gradients: $\nabla_{\theta_b} L_{\text{aux}}$, $\nabla_{\theta_a} L_{\text{aux}}$
        \STATE Update client model: $\theta_b \leftarrow \theta_b - \eta_b \nabla_{\theta_b} L_{\text{aux}}$
        \STATE Update auxiliary classifier: $\theta_a \leftarrow \theta_a - \eta_b \nabla_{\theta_a} L_{\text{aux}}$

        \STATE \textbf{// Server Side: Forward and Backward}
        \STATE \textbf{Receive} $\{\mathbf{z}_i\}_{i=1}^B$ from client
        
        \FOR{each activation $\mathbf{z}_i$ with label $y_i$}
            \STATE Compute prediction: $\hat{y}_i = M_t(\mathbf{z}_i)$
        \ENDFOR
        
        \STATE Compute global loss: $L= \frac{1}{B}\sum_{i=1}^B \mathcal{L}(\hat{y}_i, y_i)$
        \STATE Compute gradient: $\nabla_{\theta_t} L$
        \STATE Update server model: $\theta_t \leftarrow \theta_t - \eta_t \nabla_{\theta_t} L$
        
        
    \ENDFOR
\ENDFOR

\RETURN $M_b$, $M_t$, $C_a$

\end{algorithmic}
\end{algorithm}

\subsection{Training Procedure}
The training algorithm alternates between the client and server as follows (see also Algorithm~1): 

\begin{itemize}
\item \textbf{Client forward and local backward}: The client takes a batch of input data $\{(\mathbf{x}_i, y_i)\}_i^{B}$. For each $\mathbf{x}_i$, it computes $\mathbf{z}_i = M_b(\mathbf{x}_i)$ and also $\tilde{y}_i = C_a(\mathbf{z}_i)$. The client computes the auxiliary loss $L_{\text{aux}} = \frac{1}{B}\sum_i \mathcal{L}_{\text{aux}}(\tilde{y}_i, y_i)$ for the batch. It then performs a backward pass \textit{locally} to update the parameters of $M_b$ and $C_a$.
Importantly, this step does not require any communication or input from the server.  It only uses local data and labels. After computing $\mathbf{z}_i$, the client immediately sends each $\mathbf{z}_i$ to the server. Note that the client does not need to preserve $\mathbf{z}_i$ after it has both sent it to the server and used it for its own local loss computation. It could discard these activations and begin processing the next batch.
\item \textbf{Server forward and backward}: Upon receiving the batch of ${\mathbf{z}_i}$ from the client, the server proceeds with its forward pass. It computes predictions $\hat{y}_i = M_t(\mathbf{z}_i)$ for each sample and evaluates the global loss $L = \frac{1}{B}\sum_i \mathcal{L}(\hat{y}_i, y_i)$, where here $y_i$ are the true labels for the batch. The server then performs a backward pass to update its own model $M_t$, computing $\nabla_{\theta_t} L$ and updating $\theta_t$ accordingly. In this process, the server does internally compute the gradient $\frac{\partial L}{\partial \mathbf{z}}$ (the gradient of the loss with respect to the activation from the client), but \textit{this gradient is not transmitted to the client}. It is only used for the server’s weight updates. 
After updating, the server can accept the next batch of $\mathbf{z}$ from the client, making the two sides effectively decoupled aside from the one-directional activation flow.
\end{itemize}

\subsection{Discussion}
\textbf{Convergence and Performance}: An important question is whether optimizing these two separate objectives will still cause the overall network to perform well on the true task. We build upon the insights from greedy layer-wise training literature (e.g., \cite{belilovsky2020decoupled}), which indicates that if each module is trained to minimize a properly chosen local loss, the composed network can approximate the globally optimal solution. In our case, the client’s loss $\mathcal{L}_{\text{aux}}$ is aligned with the final task since $C_a$ tries to predict the same labels $y$, and the server’s loss $\mathcal{L}$ is the final task loss. We are thus optimizing a greedy surrogate objective. That is, $M_b$ is explicitly trained to produce intermediate features that are useful for predicting $y$ via $C_a$, and $M_t$ is trained to perform the final prediction given those features. 

Importantly, there is a coupling where if $C_a$ is too powerful (e.g., a deep network), $M_b$ might overfit to making $C_a$’s job easy in a way that doesn’t translate to $M_t$. To mitigate this, $C_a$ is kept small, and we can also weight the auxiliary loss relative to the global loss. In practice, one can introduce a hyperparameter $\lambda$ that scales $\mathcal{L}_{\text{aux}}$, which trains $M_b$ to minimize $L_{\text{aux}} + \lambda \cdot L$ with $M_t$ still minimizing $L$.  A small $\lambda>0$ means we occasionally send back some information from the server to the client to inform $M_b$.  However, in our decoupled SL approach, we set $\lambda=0$ to avoid any BP signals with an aim to minimize both communication and memory usage for the client that is usually resource contrained on the edge. 

\textbf{Complexity and Overhead}: The proposed method has the same time complexity per sample as conventional SL, since we similarly compute one forward and one backward per model module per sample. However, by removing the dependency, our approach potentially reduces idle times and as noted, cuts the communication volume. The communication cost per batch is one transmission of $\mathbf{z}$ from client to server. If $|Z|$ is the size of the activation,  \textbf{our approach sends roughly the half of what conventional SL would send}. The memory cost for the client includes storing activations for its own BP through $M_b$ and $C_a$. Once that is done, those can be released. In conventional SL, the client would similarly store those, but need to hold them longer waiting for the server’s gradient. On the server, memory use is slightly reduced by not storing the cut-layer gradient for sending. 
Overall, the approach is designed to be lightweight on the client side, trading off a bit additional local computation (the auxiliary head forward/backward, which is minor) for major gains in communication and memory relief.

\begin{figure}[t]
    \centering
    \includegraphics[width=0.95\linewidth]{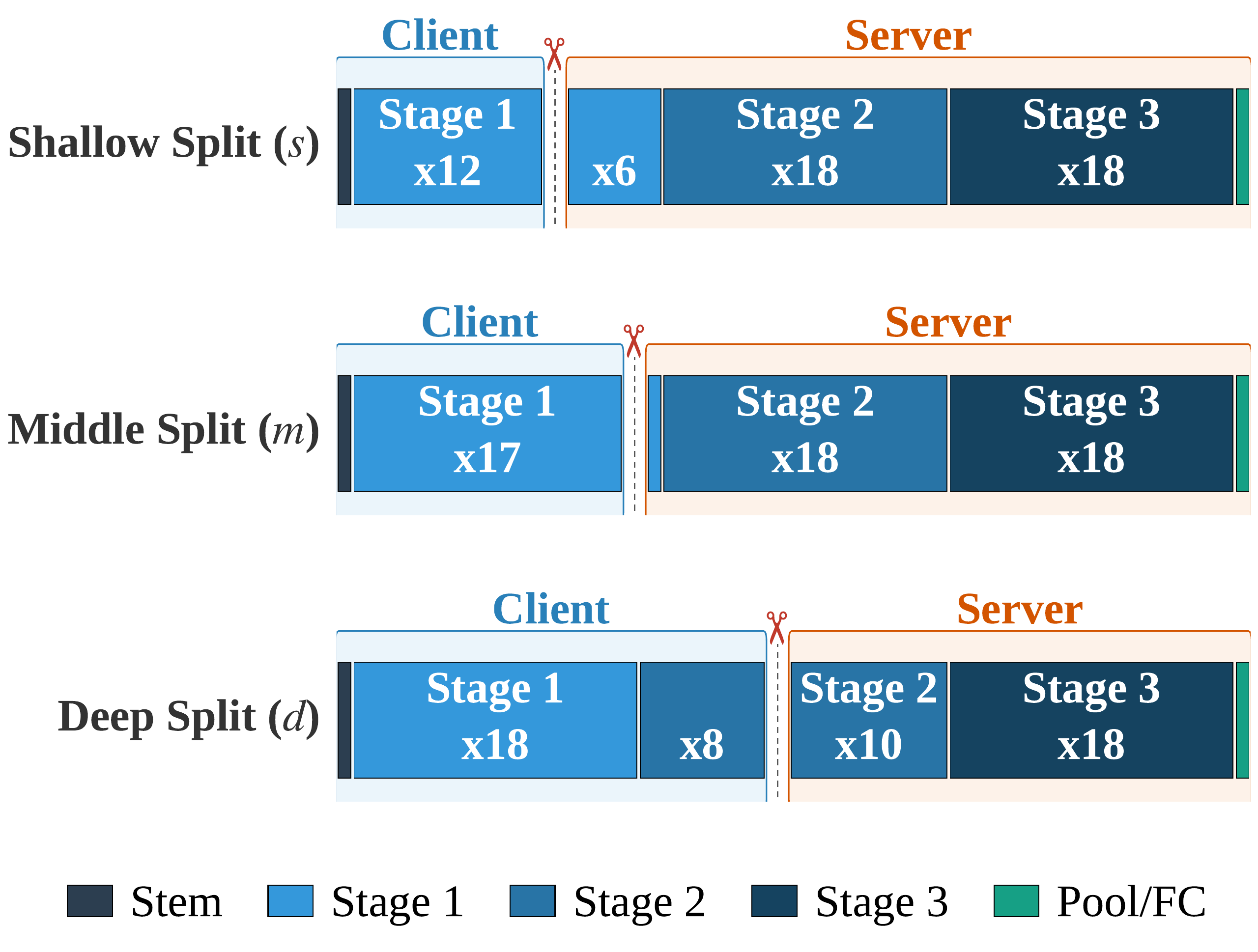}
    \caption{An illustration of three model split strategies.}
    \label{fig: cut-layer}
     \vspace{-2mm}
\end{figure}
\vspace{-1mm}
\section{Experiments}\label{sec: experiments}
\subsection{Setup} 
\textbf{Datasets and Models.} We evaluate our approach on CIFAR-10 and CIFAR-100.  \cite{krizhevsky2009learning}. We use ResNet-110 and consider three different types of model split strategies denoted by $L_c=\{s, m, d\}$, where $s, m, d$ represent $shallow, midlle, deep$ splits. In particular, ResNet-110 has in total $54$ residual blocks, and $L_c=\{s, m, d\}$ means that we keep $\{12, 17, 26\}$ blocks at the client and the remaining blocks at the server, respectively (see Figure~\ref{fig: cut-layer}).
We compare our decoupled split learning (DSL) approach to conventional SL (CSL) that uses back propagation \cite{gupta2018distributed}. For both DSL and CSL, we evaluate the performance with different client numbers $N=\{1,5,10\}$.

The key hyper-parameters are as follows. The client’s auxiliary classifier $C_a$ is a single fully-connected layer that maps the activations $\mathbf{z}$ to 10/100 classes, plus a softmax layer. The epoch number $E$ is set to be $50$. We use stochastic gradient descent with momentum $0.9$, together with a learning rate $0.001$, and a mini-batch size $128$.
\begin{figure}[t]
    \centering
    \includegraphics[width=1\linewidth]{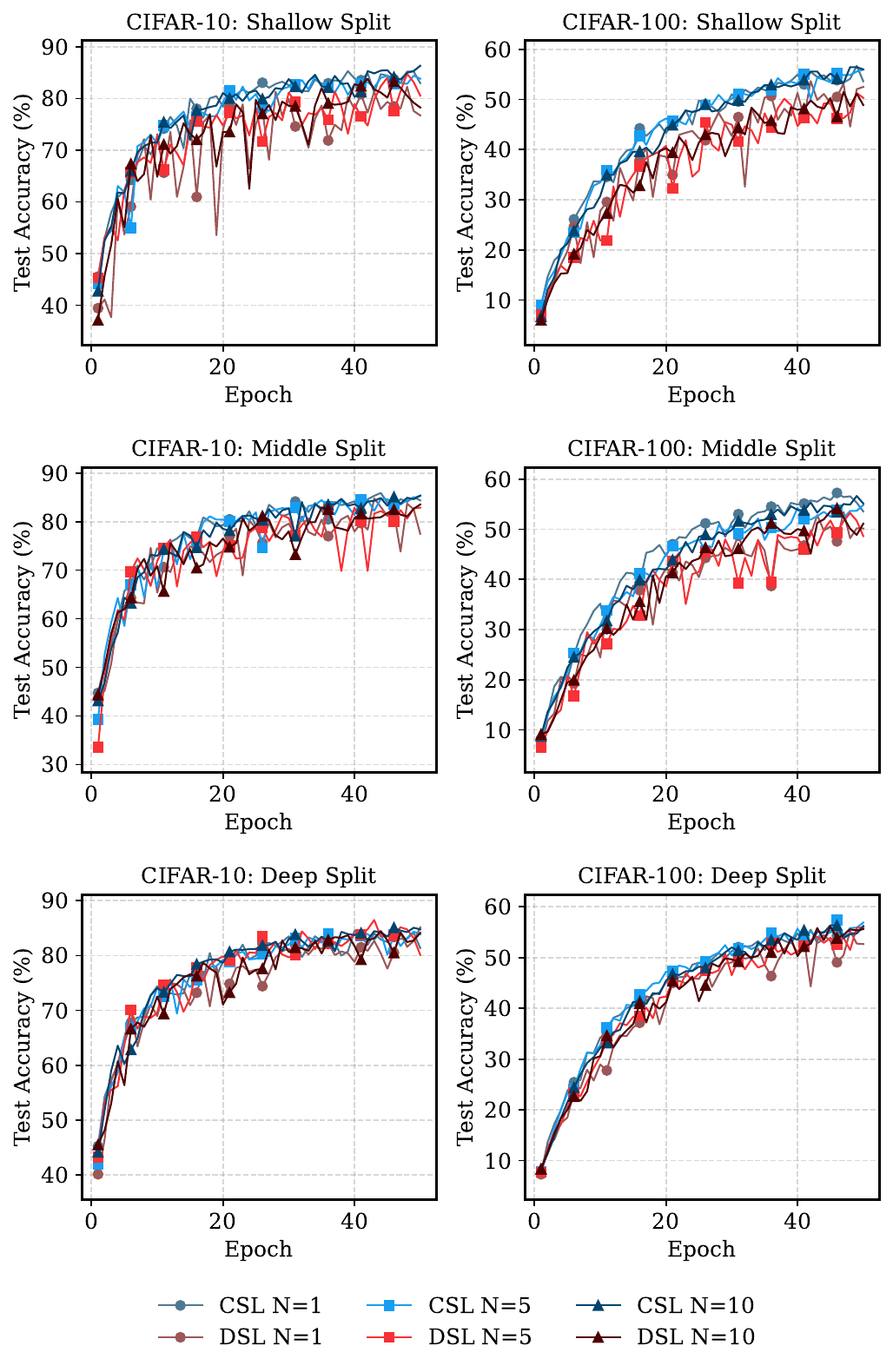  }
    \caption{Convergence of DSL and CSL under different $N$ and $L_c$.}
    \label{fig:test_acc_d}
     \vspace{-3mm}
\end{figure}


\textbf{Metrics.} We evaluate our approach using a few metrics:
\begin{enumerate}
\item \textit{Test accuracy}. We measure the convergence behavior.
\item \textit{Communication volume}. We track the communication volume exchanged between a single client and the server. For CSL, it involves both forward ($\text{Fwd}$) and backward ($\text{Bwd}$) propagation. For DSL, it only involves the forward process.
\item \textit{Memory usage}. We measure the peak memory usage 
\item \textit{Training latency}. We analyze the detailed training latency overhead in terms of forward and backward propagation as well as communication volume.  
\end{enumerate}



\subsection{Experiment Results}
\textbf{Convergence and Performance (Q1).} 
Figure~\ref{fig:test_acc_d} plots the test accuracy over training epochs. 
We observe that CSL and DSL exhibit similar convergence behavior on CIFAR-10 and CIFAR-100, both achieving accuracies that are comparable across different numbers of clients $N$ and model split strategies $L_c$. The learning curves indicate that our decoupling approach does not necessarily introduce optimization instability, and the performance gap between CSL and DSL remains small during training. 
In addition, across different model splits $L_c$, we observe that deeper cuts tend to achieve better performance (e.g., DSL with $N=5$), which is consistent with \cite{dachille2025impact}. 
\begin{Obs}
DSL achieves model performance on par with CSL based on back propagation.
\end{Obs}




\begin{figure}[t]
    \centering
    \includegraphics[width=1\linewidth]{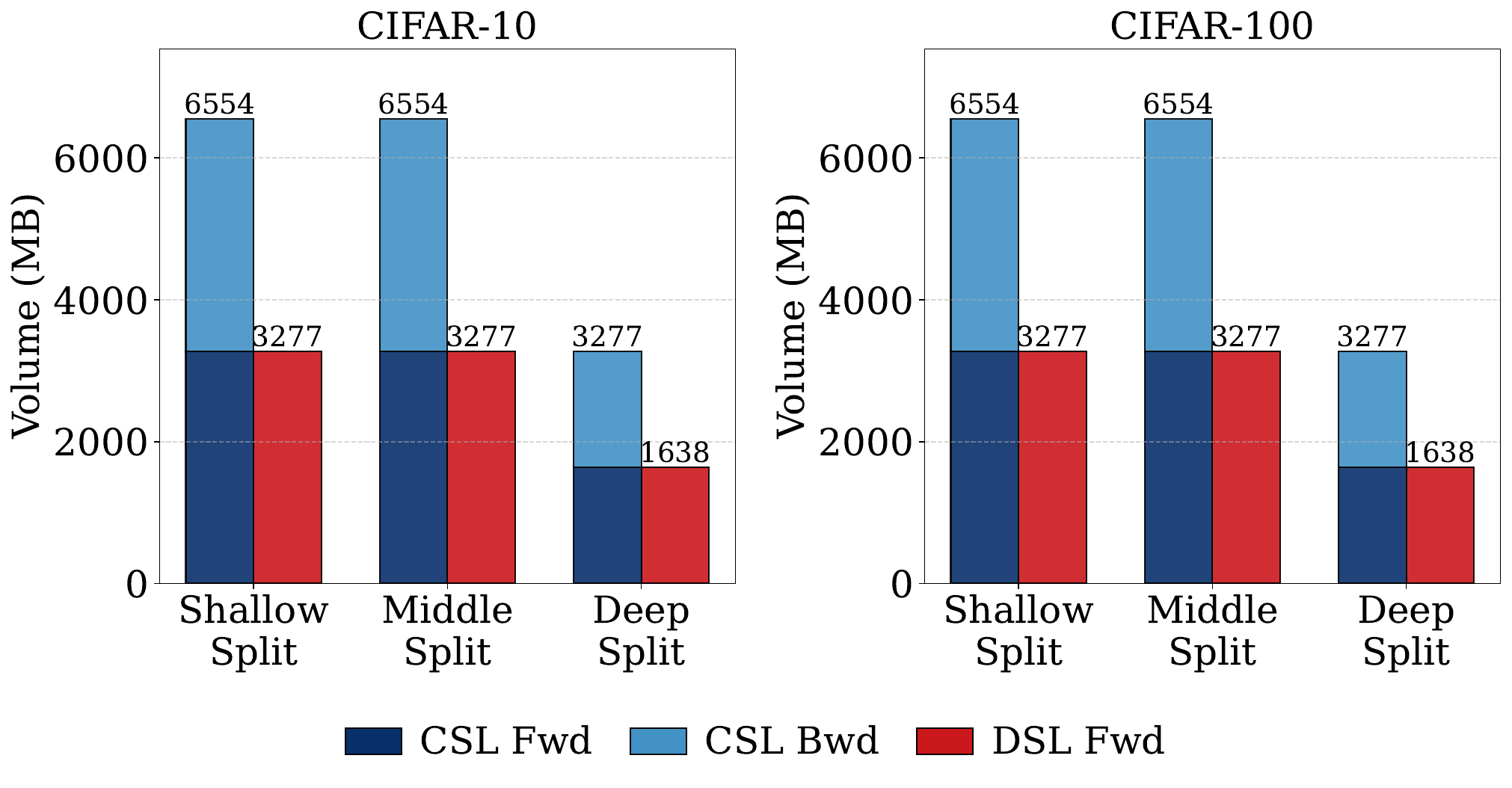}
    \caption{Communication breakdown ($\text{Fwd}$ and $\text{Bwd}$) with $N=10$.}
    \label{fig:comm_volume}
    \vspace{-2mm}
\end{figure}

\textbf{Communication (Q2).} 
Figure~\ref{fig:comm_volume} reports the communication volume breakdown between forward and backward transmissions under different model splits $L_c$ and $N=10$.\footnote{Note that results are invariant to client  number $N$, as the communication volume depends on the size of the cut layer.} When the cut layer is placed early in ResNet-110, CSL incurs substantial backward communication due to the transmission of gradients, which results in a total communication volume of approximately 6.5~GB over training on both datasets. In contrast, DSL eliminates backward communication by design, and it reduces the communication volume by nearly 50\% while preserving the same forward communication cost. As the cut layer moves toward the deep, the overall communication volume decreases for both methods due to the reduced cut layer dimensionality. 
\begin{Obs}
DSL reduces communication overhead by eliminating backward communication, and it achieves up to a 50\% reduction in communication volume compared to CSL.
\end{Obs}

\begin{figure}[t]
    \centering
    \includegraphics[width=1\linewidth]{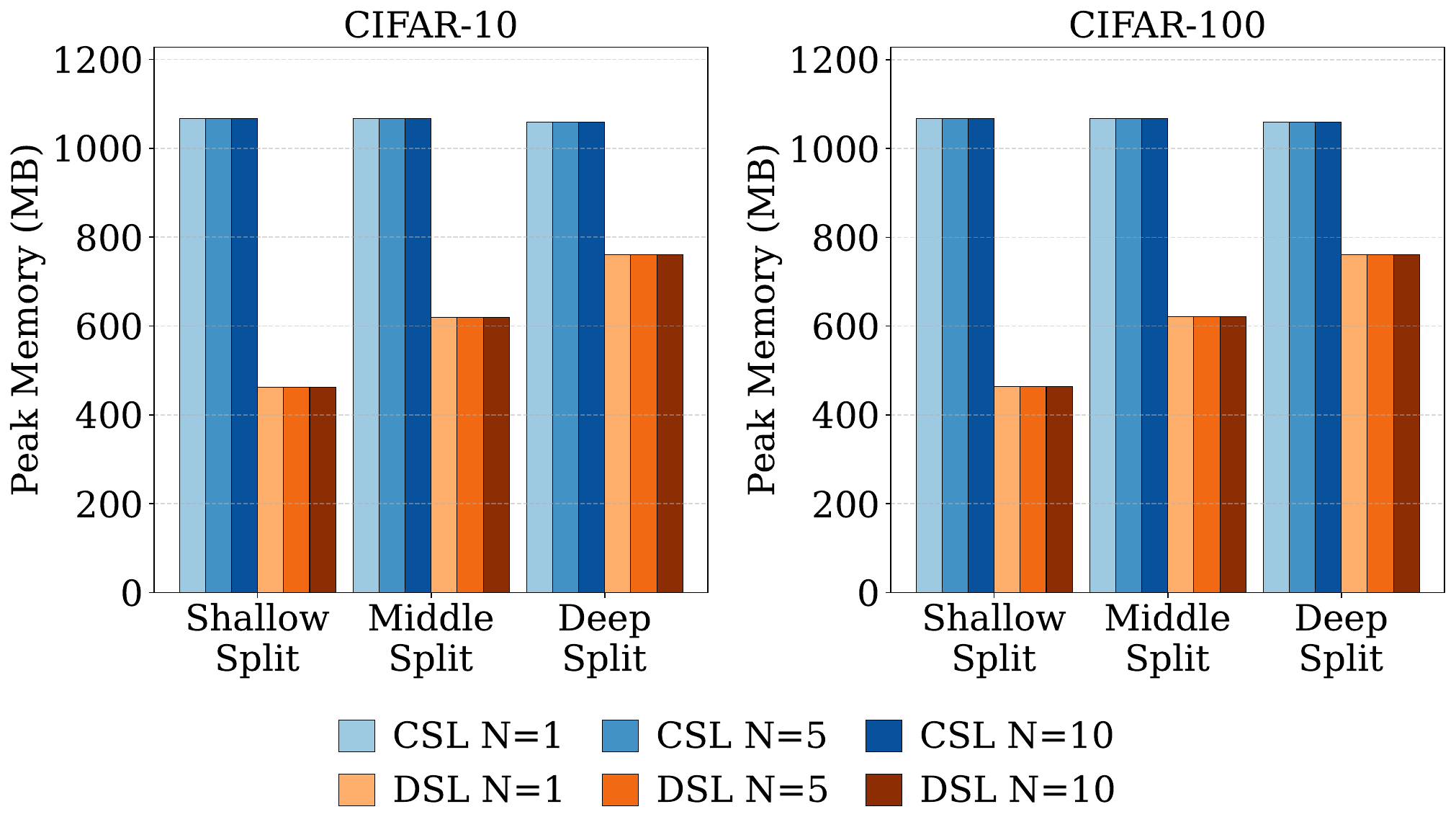}
    \caption{Peak GPU memory usage under different $N$ and $L_c$.}
    \label{fig:memory}
    \vspace{-2mm}
\end{figure}

\textbf{Memory Usage (Q2).} Figure~\ref{fig:memory} reports the peak memory usage under different client numbers $N$ and model splits $L_c$. For CSL, the peak memory footprint remains largely constant across different choices of $L_c$, as BP requires storing intermediate activations regardless of where the cut layer is placed. In contrast, DSL exhibits a clear reduction in peak memory usage as the cut layer moves shallower into the network. When $L_c=s$, DSL reduces the peak memory consumption by up to $58\%$ compared to CSL on both CIFAR-10 and CIFAR-100. This reduction stems from the elimination of activation storage needed for BP, with earlier cuts further limiting the total size of activations that must be retained at the client. The observed trends are consistent across datasets and numbers of clients, which shows that DSL provides substantial and robust memory savings without sacrificing model performance.
\begin{Obs}
DSL significantly reduces peak memory usage compared to CSL by removing the need to store activations for back propagation. The saving is larger when the cut layer is placed earlier in the neural network.
\end{Obs}

\begin{figure}[t]
    \centering
    \includegraphics[width=1\linewidth]{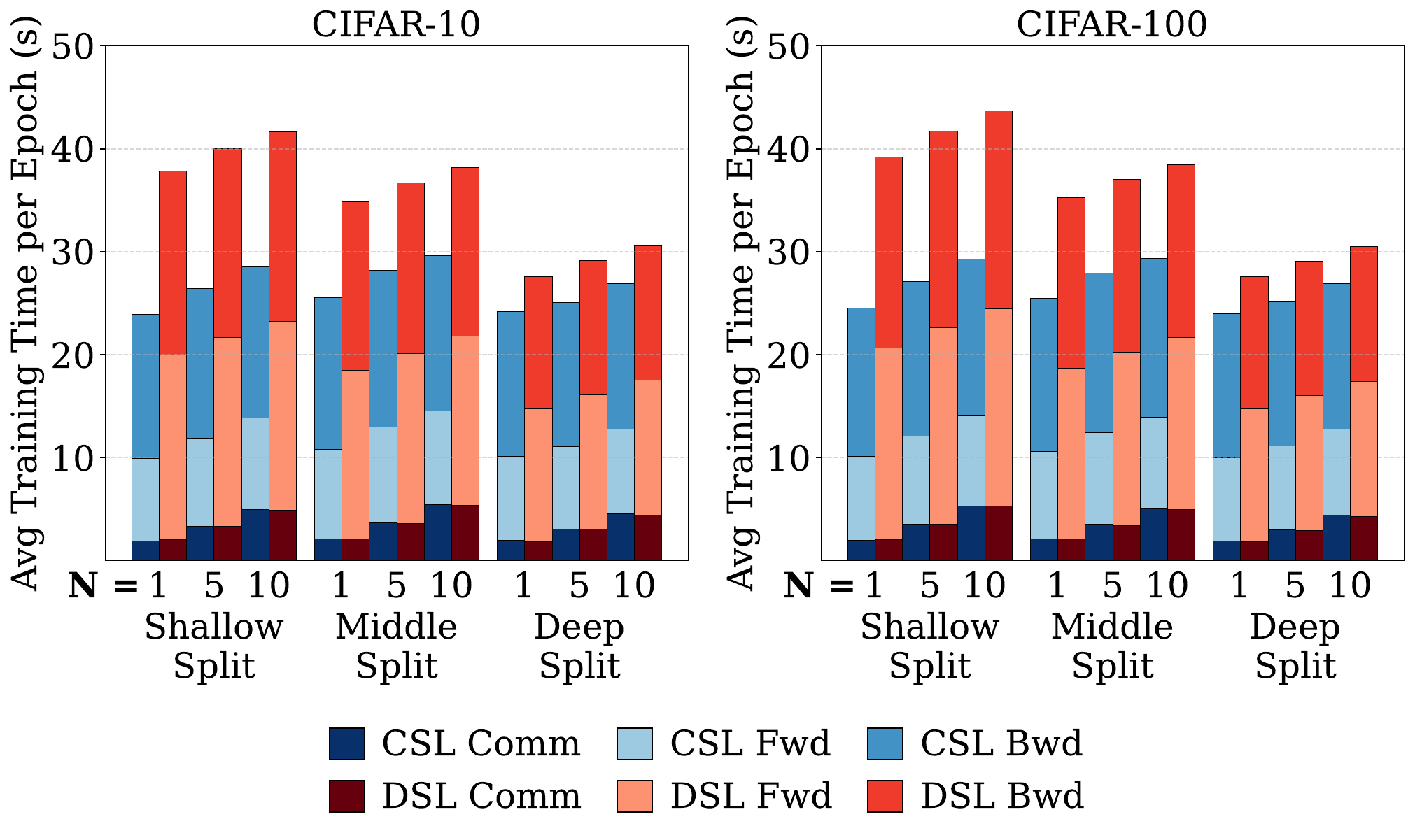}
    \caption{Training time breakdown (Fwd, Bwd and Comm).}
    \label{fig:latency}
     \vspace{-2mm}
\end{figure}

\textbf{Training Latency}. \textit{The communication and memory saving of DSL come at a cost of additional computation at the client side due to the auxiliary network.} To quantify this, 
Figure~\ref{fig:latency} reports the average per-epoch training time and decomposes it into forward computation, backward computation, and communication. Across both CIFAR-10 and CIFAR-100, we observe that DSL incurs \textit{moderately higher} total training time than CSL under most cut-layer placements, with the gap becoming more significant when the cut layer is placed shallower (e.g., $L_c=s$). The breakdown suggests that this overhead is primarily computational. That is, DSL introduces additional local optimization work due to the auxiliary network component. Meanwhile, communication contributes a small fraction of the total time for both methods and does not offset the added compute in DSL. These results indicate that DSL trades a modest increase in per-epoch runtime for the substantial communication and memory savings shown in Figures~\ref{fig:comm_volume} and~\ref{fig:memory}.
\begin{Obs}
DSL introduces a moderately higher per-epoch runtime overhead compared to CSL. 
\end{Obs}

\section{Conclusion}\label{sec: conclusion}
In this paper, we proposed a decoupled split learning framework that removes the need for backpropagation across the client–server boundary by introducing an auxiliary network at the cut layer. By allowing the client and server to optimize their respective model partitions using local objectives, the proposed approach eliminates gradient dependency and requires only one-way activation communication during training. Experiments on CIFAR-10 and CIFAR-100 demonstrate that our approach achieves model performance comparable to conventional split learning based on backpropagation. 
In addition, our approach provides substantial efficiency benefits in that it reduces total communication volume by up to 50\%, and it significantly lowers client-side peak memory usage without needing to store activations for cross-split backpropagation. 
For future work, it is interesting to design auxiliary network to reduce the computational overhead.


\bibliographystyle{ieeetr}
\bibliography{sample}

\end{document}